\documentclass[letterpaper, 10 pt, conference]{ieeeconf}

\IEEEoverridecommandlockouts                              % This command is only needed if 
\usepackage{graphicx}
\usepackage{caption}
\usepackage{subcaption}
\usepackage{amsmath,amssymb}
\usepackage[bookmarks=true]{hyperref}
\usepackage{dsfont}
\usepackage{bm}
\usepackage{bbm}
\usepackage{algorithm}
\usepackage{algorithmic}
\usepackage[algo2e, ruled, noend]{algorithm2e}

\newcount\Comments % 0 suppresses notes to selves in text
\usepackage{array,multirow,graphicx}
\usepackage{color}
\usepackage{wrapfig}
\definecolor{darkgreen}{rgb}{0,0.5,0}
\definecolor{purple}{rgb}{1,0,1}
\newcommand{\kibitz}[2]{\ifnum\Comments=1\textcolor{#1}{#2}\fi}
\DeclareMathOperator*{\argmin}{arg\,min}
\DeclareMathOperator*{\argmax}{arg\,max}
\usepackage{xcolor}

\title{\LARGE \bf
Maximizing BCI Human Feedback using Active Learning
}

\author{
Zizhao Wang$^{*}$, Junyao Shi$^{*}$, Iretiayo Akinola$^{*}$ and Peter Allen
\thanks{$^*$Equal Contribution.}
\thanks{ Department of Computer Science, Columbia University, New York. This work was supported in part by a Google Research grant and National Science Foundation grant IIS-1527747.
} 
}

\begin{document}

\maketitle
\thispagestyle{empty}
\pagestyle{empty}

%%%%%%%%%%%%%%%%%%%%%%%%%%%%%%%%%%%%%%%%%%%%%%%%%%%%%%%%%%%%%%%%%%%%%%%%%%%%%%%%
\begin{abstract}
Recent advancements in \textit{Learning from Human Feedback} present an effective way to train robot agents via inputs from non-expert humans, without a need for a specially designed reward function. 
However, this approach needs a human to be present and attentive during robot learning to provide evaluative feedback. In addition, the amount of feedback needed grows with the level of task difficulty and the quality of human feedback might decrease over time because of fatigue.
To overcome these limitations and enable learning more robot tasks with higher complexities, there is a need to maximize the quality of expensive feedback received and reduce the amount of human cognitive involvement required.
In this work, we present an approach that uses active learning to smartly choose queries for the human supervisor based on the uncertainty of the robot and effectively reduces the amount of feedback needed to learn a given task. We also use a novel multiple buffer system to improve robustness to feedback noise and guard against catastrophic forgetting as the robot learning evolves. This makes it possible to learn tasks with more complexity using lesser amounts of human feedback compared to previous methods. We demonstrate the utility  of our proposed method on a robot arm reaching task where the robot learns to reach a location in 3D without colliding with obstacles. Our approach is able to learn this task faster, with less human feedback and cognitive involvement, compared to previous methods that do not use active learning.
\end{abstract}

%%%%%%%%%%%%%%%%%%%%%%%%%%%%%%%%%%%%%%%%%%%%%%%%%%%%%%%%%%%%%%%%%%%%%%%%%%%%%%%%

\section{Introduction}
Learning from human feedback (LfHF) is an effective way to teach robot agents new skills. In this learning paradigm, an artificial agent receives feedback signals from a human expert that is watching the agent learn  \cite{knox2009interactively}\cite{griffith2013policy}\cite{christiano2017deep}\cite{warnell2018deep}.
However, LfHF requires a human to be present during the learning process to provide the evaluative feedback. Depending on the difficulty of the task, the learning process might require a large amount of feedback to effectively learn the task, which translates to a significant amount of human time.

Human feedback can be collected via a variety of means including mouse clicks~\cite{christiano2017deep}\cite{jagodnik2017training}, facial expressions \cite{veeriah2016face}, finger pointing \cite{cruz2016multi} and via human physiological signals like brain signals \cite{iturrate2015teaching}\cite{akinola2019accelerated} among others. Learning from physiological signals is appealing in that the human does not need to perform any extra actions like mouse clicking etc, besides watching and evaluating the robot. On the other hand, learning from measured physiological signals such as brain signals measured via electroencephelography (EEG) presents a unique challenge; the physics of the EEG devices limit their signal-to-noise ratio which consequently results in noisy decoding of the evaluative feedback.
More feedback might be needed to overcome increased noise levels in it; this solution is undesirable due to its high time cost.

\begin{figure}[t]
\vspace{2mm}
\begin{center}
    \begin{subfigure}[h]{1\linewidth}
        \centering
        \includegraphics[width=0.6916\linewidth]{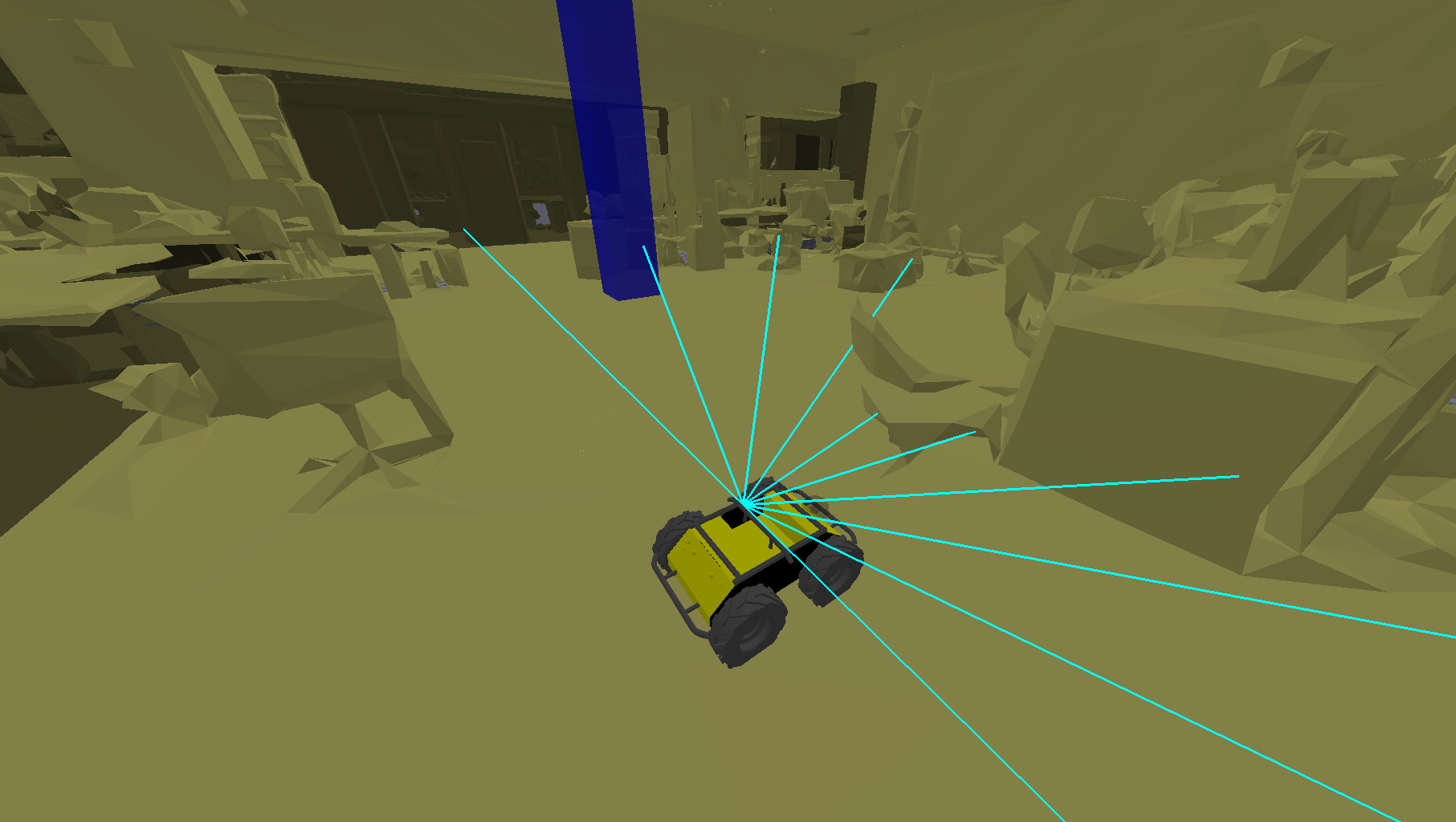}
    \end{subfigure}
    \par\smallskip % \smallskip , \medskip and \bigskip
    \begin{subfigure}[h]{1\linewidth}
        \centering
        \includegraphics[width=0.7125\linewidth]{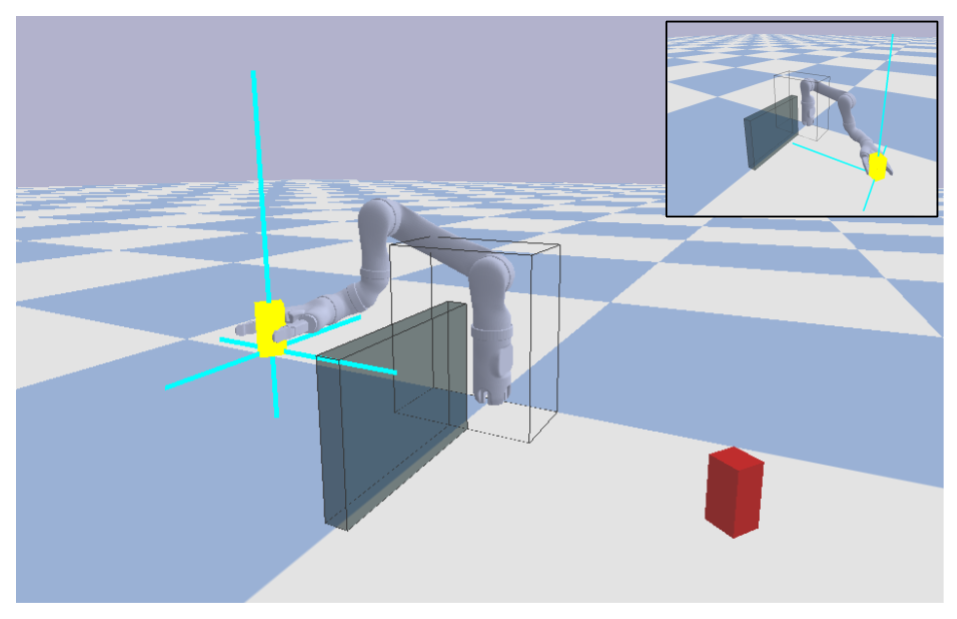}
    \end{subfigure}
\end{center}
\setlength{\abovecaptionskip}{-0pt}
\setlength{\belowcaptionskip}{-21pt}
\caption{\small Task environments. \textbf{Top}: For the navigation task, the robot agent learns an obstacle-avoidance policy to reach the goal (blue). The agent senses its environment with laser scans (10 cyan rays). \textbf{Bottom}: In the reaching task, the robot arm moves its end-effector to place the yellow block at the goal (red). The challenge is to navigate around the obstacle (slightly transparent black box) to reach the goal. Inset is an example of successful reach. For both tasks, in the sparse reward RL setting, the agent is unable to avoid obstacles and reach the goal. Compared with the navigation task, the large 3D state space in the reaching task would require a large amount of human feedback. Our proposed method reduces feedback required and is able to learn this task using evaluative feedback decoded from human EEG.}
\label{fig:reaching_task}
\vspace{3mm}
\end{figure}

In this work, we investigate ways to reduce the amount of human feedback decoded from physiological signals needed to teach robot agents new tasks. This would reduce the amount of human hours required to teach the robot and also enable learning more complex tasks that previously required a large amount of feedback.
We propose an active-learning-like approach that selects which queries to ask the human rather than ask the user at every move at every time-step. While a simulated robot can move very fast during the learning process, we optimally choose when to slow the robot down to get human feedback. Otherwise, the robot is able to speedily step through many steps when not asking for feedback. We use a special replay buffer to keep consensus feedback data to enable robustness to noise and guard against catastrophic forgetting.

Experiments using simulated feedback signals from a noisy oracle show that our algorithm is able to learn a complex reaching task (with a large state and action space) in a feedback-efficient way. A baseline method would require a larger amount of feedback and a prohibitively long amount of time to learn the task from real human feedback.
In summary, our method addresses the challenge of robot learning from limited (in quantity and quality) human feedback. Short human attention span and the imperfect signal-to-noise ratio of many brain-computer-interface devices restrict the quantity and quality of the evaluative feedback obtained from human physiological signals (such as brain signals). We propose to alleviate some of the challenges algorithmically.
Our main contributions include:
\begin{itemize}
    \item We introduce an active query approach that models a robot's decision uncertainty with a Dirichlet distribution and queries the human for feedback only when the robot is unsure.
    \item We present the novel use of a \textit{purified} buffer to enable robustness to highly noisy human feedback.
    \item We demonstrate the application of our method to a challenging task which has large state and action spaces and is otherwise difficult to learn from small amount of noisy feedback.
\end{itemize}

\section{Related Work}

\subsection{Learning from Human Feedback}

Significant research effort has been directed toward learning from human feedback (LfHF) and its combination with reinforcement learning (RL) to teach robots different skills \cite{knox2009interactively}\cite{griffith2013policy}\cite{christiano2017deep}\cite{warnell2018deep}.  A natural approach is to use feedback as the reward signal to train a reinforcement learning agent, such as TAMER \cite{knox2009interactively}\cite{warnell2018deep}. A potential drawback of using feedback purely for RL is that inconsistent feedback can yield suboptimal performance. This can be addressed by combining human feedback with hand-crafted markov decision process (MDP) reward functions \cite{knox2012reinforcement}\cite{knox2010combining}. In a two-stage process, the policy first learned from human feedback is then fine-tuned by RL. This assumes that the learned feedback policy from the first stage provides good initialization for RL. Although we  also use this two stage approach, our work is orthogonal in that we focus on reducing the amount of feedback needed to have a good LfHF policy to guide RL stage.

Another approach to learning from feedback is to derive a reward function from the feedback \cite{christiano2017deep} rather than using the feedback directly as rewards. The derived reward function is then used for RL. In that work, human feedback is provided in the form of preferences between pairs of trajectory segments indicated by mouse clicks. In contrast, we focus on learning from physiological signals that do not require mouse clicks and obtain human evaluative feedback for each state-action pair chosen by the agent. We use active learning to reduce the amount of feedback needed in this setting.

\subsection{Active Learning}
According to a survey\cite{settles2009active}, active learning is a learning algorithm that can attain superior performance with fewer data by selecting which data to learn from. This approach selects data points that are optimal with respect to some information-seeking criteria in a way that minimizes the total number of queries required. Several works have used this technique in machine learning settings \cite{agarwal2013selective}\cite{orabona2011better}\cite{dekel2006online} and especially for robotics \cite{cakmak2012designing}\cite{hangl2017skill}\cite{kulick2013active}.
We use this key insight of active learning to reduce the amount of required feedback and enable learning hard tasks directly from the brain. Asides from our distinct application, we differ from existing works in the idea of selectively choosing to slow down some robot actions for the user to observe while the robot speeds through other actions during training. In addition, we use a Dirichlet distribution to measure robot uncertainty and actively choose actions with high uncertainties.

% \todo{TODO: Add a synopsis of our approach and why it is novel- add contributions.  Mention  (and describe) our old work and how this extends it}

\subsection{Brain-Computer Interface (BCI) Robot Learning}

Error-related potentials (ErrPs) occur in the human brain when a human observes an error during a task \cite{spuler2015error} and a few works have explored using this as evaluative feedback to train robot in the context of robot learning.
For example, \cite{schiatti2018human} used ErrPs as reward signals in the reinforcement learning setting to teach a robot a reaching task. The task was limited to 2D task by limiting the points to a plane of 5x5 possible points. In contrast, the reaching task in this work is 3D with an obstacle that the robot has to learn to avoid.
More recently, in our previous work \cite{akinola2019accelerated}, we developed an algorithm to use evaluative feedback from human brain to enhance learning a  navigation task in a sparse reward setting. In that work, we trained a supervised learning model using noisy evaluative feedback from decoded ErrP signals. The obtained supervised policy is suboptimal but provides useful coarse guidance for the subsequent RL learning stage that uses a sparse reward to achieve an optimal policy. 
Now, we build on our previous work and address the unique issues of learning from brain signals. These include dealing with inherent noisiness of the feedback obtained via BCI devices and reducing the overall amount of feedback needed to learn a task. Our proposed solution enables learning more complex tasks with higher dimensional state and action spaces.

\begin{figure}%[t]
\vspace{2mm}
\begin{center}
    \includegraphics[width=0.8\linewidth]{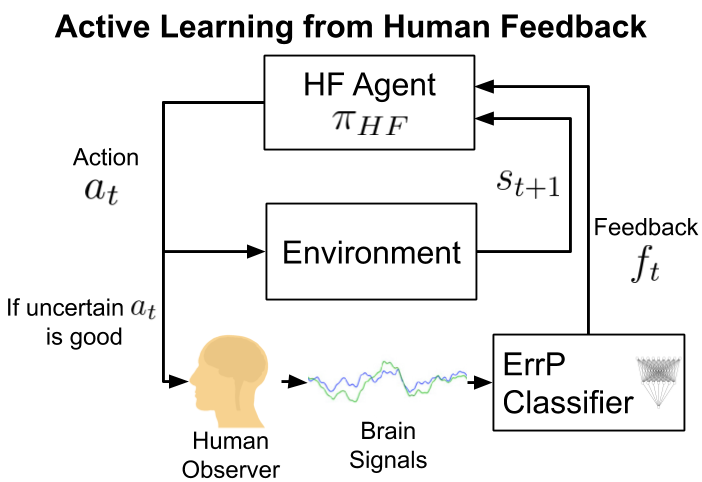}
\end{center}
\setlength{\belowcaptionskip}{-15pt}
\caption{\small Active LfHF. Instead of query labels for every action, we only ask for feedback that $\pi_{HF}$ is unsure to be good. This improves information gain from each query and boosts feedback-efficiency.
}
\label{fig:system_fig}
% \vspace{-3mm}
\end{figure}

\section{Method}

There are three stages in our Active Brain-Guided RL algorithm: train an EEG classifier so that we can infer the human feedback from brain signals; learn a Human Feedback (HF) policy based on the actively queried feedback; improve RL policy learning from sparse rewards by learning to reproduce good decisions generated by the HF policy.

\subsection{EEG Classifier Training}

Similar to \cite{akinola2019accelerated}, to evaluate human feedback from EEG signals, we collect EEG data and corresponding ErrP labels in an offline session. Then with collected data, we train a classifier, denoted as $g(\cdot; \theta_{EEG})$, which enables us to detect ErrPs when the human observes the robot choose wrong actions.
In detail, first, the human subject gets familiar with the paradigm in a short training. Then the subject will watch the robot conducting a random policy while EEG signals and the judgments of actions (good/bad) are simultaneously recorded. 
In our reaching task paradigm, we obtain ground truth judgments as labels with the Dijkstra algorithm. For tasks where ground truth cannot be easily scripted, a human expert or the subject can provide the labels. 
In our experiments, the EEG signals are recorded at 2048 Hz using 64 channels of the BioSemi EEG Headset and around 600 data points are collected.

After data collection, we preprocess EEG data and train the classifier. We filter EEG data to 1-30 Hz, apply baseline correction with $[-0.2, 0]$s before stimuli as the baseline interval, and extract EEG trails at $[0, 0.8]$s after the robot takes actions. Then we train the classifier modified from EEGNet \cite{lawhern2018eegnet} using the cross-entropy loss $\mathcal{L}_{EEG}$, where detailed architecture modifications are described in \cite{akinola2019accelerated}. During experiments, the testing accuracies among different subjects range between $0.55 - 0.73$.

\subsection{Human Feedback Policy}

With the trained EEG classifier, when subjects observe the robot's action $a_t$ at state $s_t$, we can determine the corresponding feedback $f_t$ from their brain signals. Note that in our work $f_t$ can be quite noisy as the EEG classifier has low accuracy. To learn a human feedback policy, we assume the subject first forms a policy she/he thinks is optimal and then judges if $a_t$ is good by checking if $a_t$ is consistent with her/his optimal policy at $s_t$. Formally, we denote this judgement process as a function $F: S \times A \rightarrow [0, 1]$, where $0/1$ denotes $a_t$ is bad/good respectively. During an online session, we collect feedback data $(s_t, a_t, f_t)$ in a replay buffer, and simultaneously we learn an approximator of $F$ which is denoted as $\hat{F}$ and derive the HF policy as:
\begin{equation}
\pi_{HF}(s) = \argmax_{a}\hat{F}(s, a)
\end{equation}

In our work, we use a fully connected network as the approximator and $\hat{F}$ is continually trained with the feedback pairs $(s_t, a_t, f_t)$ generated by $\pi_{HF}$.
Despite the straightforward setup, there are two challenges to learn a good HF policy: (1) the limited amount of human feedback ($\le 1000$ labels) to learn a long-horizon task (2) the highly inconsistent feedback due to the low accuracy $(0.55-0.73)$ of the EEG classifier. To overcome these challenges, beyond having a light network (one hidden layer of 32 units) to reduce parameters to learn, we adopt the following two strategies:

\subsubsection{Active Feedback Query}
In our previous work~\cite{akinola2019accelerated}, the agent asks for feedback on every action it takes, but this wastes feedback, especially when $\hat{F}$ is well trained for the given $(s_t, a_t)$. 
Instead, as shown in Fig \ref{fig:system_fig}, the agent measures its confidence that $a_t$ is good and then only query feedback if it is not confident enough (less than the threshold $\epsilon_{AQ}$).
Inspired by active learning, this strategy enables the agent to learn most from each query and improves feedback efficiency.
However, we need to notice learning the HF policy is still different from active learning scenarios since it is solving a sequential decision making problem rather than classifying data points.
In detail, we adopt Evidential Deep Learning (EDL)~\cite{sensoy2018evidential} for $\hat{F}$. Rather than output a deterministic probability for $a_t$ to be good, $\hat{F}$ generates a Dirichlet distribution over all such probabilities from which we can measure the prediction confidence.
Given an input $s_t$, $\hat{F}$ outputs the evidences that each possible action $a_i$ is either good (denoted as $e_{ig} \ge 0$) or bad ($e_{ib} \ge 0$).
Then the Dirichlet distribution $D_i$ over all possible classification probability, $\bm{p_i} = [p_{ig}, p_{ib}]$, is formed with the parameter $\bm{\alpha_i} = [\alpha_{ig}, \alpha_{ib}] = [e_{ig} + 1, e_{ib} + 1]$ as
\begin{equation}
D_i(\bm{p_i}|\bm{\alpha_i}) = \frac{1}{B(\bm{\alpha_i})} p_{ig}^{\alpha_{ig} - 1} p_{ib}^{\alpha_{ib} - 1},
\end{equation}
where $B(\bm{\alpha_i})$ is two-dimensional beta function and $p_{ig}, p_{ib}$ are the probabilities that action $a_i$ is good or bad respectively.

Note the dependence of the generated distribution $D_i$ on the state since its parameter $\bm{\alpha_i}$ is a function of the evidences which depend on the state via $\hat{F}$.
For a given state, the probability that each action $a_i$ is good is evaluated as the mean of the Dirichlet distribution: $\hat{p}_{ig} = \frac{\alpha_{ig}}{\alpha_{ig} + \alpha_{ib}}$. The confidence of each prediction is measured as the difference between the generated $D_i$ and the Dirichlet distribution having uniform probability density, i.e. the one with parameters $\bm{\alpha} = [1, 1]$.
The intuition is that the uniform Dirichlet distribution represents high uncertainty, and a more confident Dirichlet distribution $D_i$ would be further from uniform in the parameter space. This distance in the parameter space (confidence) is measured and normalized to $[0, 1)$ as:
\begin{equation}
c_i = \frac{\alpha_{ig} + \alpha_{ib} - 1 - 1}{\alpha_{ig} + \alpha_{ib}} = \frac{e_{ig} + e_{ib}}{e_{ig} + e_{ib} + 2}.
\end{equation}
Finally, the parameters are learned by minimizing the expectation of cross-entropy loss, with KL divergence from the uniform Dirichlet distribution as the regularization
\begin{align}
\mathcal{L} =& \int -(\mathds{1}_{f_t = 0}\log(p_{ib}) + \mathds{1}_{f_t = 1}\log(p_{ig})) D_i({p}_i|\bm{\alpha}_i) d\bm{p}_i \nonumber\\
&+ \lambda KL\left[D_i(\cdot|\bm{\alpha}_i) \parallel D(\cdot | [1, 1])\right],
\end{align}
where $\lambda$ is the regularizing coefficient increases from 0 to 1.
% where $\lambda_t = \min(1.0, t/10)$ is the annealing coefficient.

We also tried Gaussian Process as the predictor which naturally provides both prediction and confidence measures. However, it easily overfits the inconsistent feedback and is much more computationally expensive than neural networks.

\subsubsection{Purified Buffer (PB)} 
Beyond improving the feedback efficiency, the confidence measure from EDL also provides a tool to mitigate the significant inconsistency in the feedback.
The key observation is simple: after training, $\hat{F}$ should fit consistent feedback data better than inconsistent ones, as consistent ones still take up the majority despite low feedback accuracy.
In other words, for feedback pairs $(s_t, a_t, f_t)$ that are consistent, $\hat{F}(\cdot|s_t, a_t)$ should have the same prediction as the label $f_t$, as well as good confidence measures $c_t \ge \epsilon_{PB}$, where $\epsilon_{PB}$ is the confidence threshold.
Hence, apart from the main replay buffer collecting all $(s_t, a_t, f_t)$ pairs, we hold a second ''purified" buffer to store the feedback pairs satisfying this condition.
In this way, it will contain feedback of much higher accuracy and can stabilize the $\pi_{HF}$ learning against feedback noise.
As a result, every time we train $\hat{F}$, we will sample batches from both normal replay buffer and the purified buffer.
Besides, the purified buffer also helps the agent review important past feedback and reduces the chance of catastrophic forgetting.

\begin{algorithm2e}%[H]
\SetAlgoNoLine
\caption{\strut Active Brain-Guided RL}
\KwData{offline EEG signals $x_{1:M}$ and labels $f_{1:M}$}
\textbf{Train the EEG classifier.}
\begin{equation*}
\theta_{EEG}^* = \argmin_{\theta_{EEG}} \frac{1}{M} \sum_{i=1}^M [\mathcal{L}_{EEG}(g(x_i; \theta_{EEG}), f_i)].
\end{equation*}

\textbf{Train the HF policy.} \\
initialize the feedback replay buffer $B_{HF} = \emptyset$. \\
\While{not received $K$ feedback labels}{
    execute an action $s_t, a_t, c_t \sim \pi_{HF}(a_t|s_t)$.\\
    \If{$c_t \ge \epsilon_{AQ}$}{
        do not query the feedback.\\
        continue to execute the next action\; 
    }
    query feedback by classifying EEG signal $f_t = g(x_t; \theta_{EEG}^*)$.\\
    construct the purified buffer $B_{PB} = \{(s_t, a_t, f_t) \in B_{HF} | \hat{F}(s_t, a_t) = f_t, c_t > \epsilon_{PB}\}$. \\
    add $(s_t, a_t, f_t)$ to $B_{HF}$. \\
    update $\hat{F}$ with batches from $B_{HF}$ and $B_{PB}$.
}
\textbf{Train the RL policy.} \\
replay and episode buffer $B_{RL} = \emptyset, B_{EP} = \emptyset$. \\
\For{$t = 1, \dots, t_{RL}$}{
    \eIf{$t \le 0.2 \cdot t_{RL}$}{
        execute an action $s_t, a_t, r_t \sim \pi_{HF}(a_t|s_t)$
    }{
        execute an action $s_t, a_t, r_t \sim \pi_{RL}(a_t|s_t)$
    }
    % execute an action $s_t, a_t, r_t$ using $\pi_{HF}$ if $t \le 0.2 \cdot t_{RL}$ else using $\pi_{RL}$\\
	add $(s_t, a_t, r_t)$ to $B_{EP}$. \\
	\If{$s_t$ is the end of the episode}{
        add $(s_t, a_t, R_t = \sum_{k=t}^\infty \gamma^{k-t} r_k)$ to $B_{RL}$ for $t$ in $B_{EP}$. \\
        clear the episode buffer $B_{EP} = \emptyset$.
    }
    optimize $\mathcal{L}^{PPO}$ with on-policy samples. \\
    optimize $\mathcal{L}^{imit}$ with batches from $B_{RL}$.
}
\label{alg:brain_guided_rl}
\end{algorithm2e}
After learning with human feedback, the policy $\pi_{HF}$ has a rough knowledge about which actions are good/bad. In spite of the low success rate to finish the task, this HF policy still provides better exploration than random, maximum entropy exploration which most RL algorithms use.

\subsection{Sparse-Reward RL with Guided Exploration and Imitation Learning}
The final stage is to learn an RL policy $\pi_{RL}$ efficiently in an environment with sparse rewards.
Typical exploration strategies struggle to stumble on positive rewards that provide learning signals.
To address this, we use $\pi_{HF}$ as the initial behavior policy during RL learning. Even though $\pi_{HF}$ may be far from perfect, it guides the exploration towards the goal and increases the chances of getting positive rewards.
In addition to learning from rewards, we let our RL agent reproduce previous good decisions (by $\pi_{HF}$ or $\pi_{RL}$), using imitation learning~\cite{oh2018self}.
To do this, we store past episodes and their cumulative rewards $(s_t, a_t, R_t = \sum_{k=t}^\infty \gamma^{k-t} r_k)$ in a replay buffer.
We use a filtered imitation learning loss that selectively learns from good experiences only. In the actor-critic framework, the loss has two components :
\begin{align}
\mathcal{L}^{imit} &= \mathcal{L}^{imit}_{actor} + \mathcal{L}^{imit}_{critic} \nonumber\\
\mathcal{L}^{imit}_{actor} &= \left[ -\log \pi_{RL}(a_t|s_t) (R_t - V(s_t)) \right] \cdot \mathds{1}_{R_t > V(s_t)} \\
\mathcal{L}^{imit}_{critic} &= \frac{1}{2} \Vert R_t - V(s_t) \Vert^2 \cdot \mathds{1}_{R_t > V(s_t)}.
\end{align}
With the filter $\mathds{1}_{R_t > V(s_t)}$, the RL agents will only learn to choose $a_t$ chosen in the past at $s_t$ if return $R_t$ is greater than current value estimate $(R_t > V(s_t))$. This enables the RL agent to focus on successful episodes from $\pi_{HF}$ in the initial training stage. Moreover, when $\pi_{RL}$ is close to convergence, the filter guarantees that $\pi_{RL}$ will not be constrained to the suboptimal performance of $\pi_{HF}$ and can outperform it.

Implementation-wise, we can choose any off-policy actor-critic RL algorithm. Our method even can be applied to on-policy Deep RL algorithms like PPO \cite{schulman2017proximal} which we adopt as $\pi_{RL}$. The policy and value networks have the same architecture as $\pi_{HF}$, except for the output layers as they have different output dimensions. For the first $20\%$ of training, we choose $\pi_{HF}$ as the behavior policy to generate good episodes for imitation learning. Then we switch to the $\pi_{RL}$.
Our full algorthim for Active Brain-Guided RL is given in Alg \ref{alg:brain_guided_rl}.

\section{Experiments}
To test our method, we implemented navigation and reaching tasks in Gibson \cite{xiazamirhe2018gibsonenv} and pybullet  environment respectively.
For the navigation task, the goal is to drive the robot (shown in Fig \ref{fig:reaching_task}, top) to the fixed goal represented by the blue pillar. The environment is a $11 \times 12 m^2$ area with multiple obstacles.
The state space is chosen as $s_t = (l_t, d_t, \phi_t) \in \mathbb{R}^{13}$ where $l_t \in \mathbb{R}^{10}$ is laser range observations evenly spaced from $90^{\circ}$ left to $90^{\circ}$ right of the robot, $d_t \in \mathbb{R}^{2}$ is the distance and relative angle from goal, and $\phi_t$ is the robot yaw angle.
The action space $A$ is discretized to help human subject identify actions and determine their optimality. Three actions: moving $0.3m$ forward, turning $30^\circ$ left or right.

For the reaching task, the goal is to control the 6-DOF Mico Robotic arm (shown in Fig \ref{fig:reaching_task}, bottom) to bring the yellow object to the fixed goal location shown as the red block. To this end, the arm needs to get around the grey obstacle while avoiding self-collision (indicated by the transparent box). The arm moves in a $0.5 \times 1.0 \times 0.675m^3$ space, which covers most reachable space for the arm.
The state space is chosen as $s_t = (p_t, l_t) \in \mathbb{R}^{9}$ where $p_t \in \mathbb{R}^{3}$ is the Cartersian coordinate of the end effector and $l_t \in \mathbb{R}^{6}$ is laser range observations along three coordinate axes.
The action space $A$ contains six actions: moving along both directions of the x/y/z-axis with constant step lengths of $0.025$/$0.05$/$0.0675m$ respectively. In this case, the state space is split to a $21 \times 21 \times 11$ grid.

For both tasks, the sparse reward \textbf{RL sparse} is -0.05 for every step, except for a +10 bonus when reaching the goal or -5 penalty for collision.
Instead, a reward function providing richer learning signals, \textbf{RL rich}, can be:
\[
    r_{rich}(s_t) = c_d \cdot d_t + c_\theta \cdot \theta_t  \qquad \text{   if no collision occurs }
\]
where $d_t$ is the distance from the goal, $\theta_t$ is the difference between the current orientation and the orientation to the goal (only for the navigation task), and $c_d = -0.01$ (and $c_\theta = -0.003$) are the empirically selected hyperparameters. This rich reward motivates the robot to get close to (and head towards) the goal, which leads to a more efficient exploration.

For the navigation task, the robot's initial position is uniformly sampled in a $0.2 \times 0.2m^2$ area.
For the reaching task, the beginning location of the end-effector is sampled in a $0.2 \times 0.1 \times 0.125m^3$ space, which is a $9 \times 3 \times 3$ grid. For both navigation and reaching tasks, depending on the initial position, the optimal path consists of $17-19$ and $33-41$ steps respectively.
The episode ends if any of the following happens: the goal is reached, a collision occurs with surrounding obstacles or the maximum  episode length of 120/80 steps is reached for the navigation/reaching tasks.

\begin{figure}%[t]
\vspace{2mm}
\begin{center}
    \includegraphics[width=1.0\linewidth]{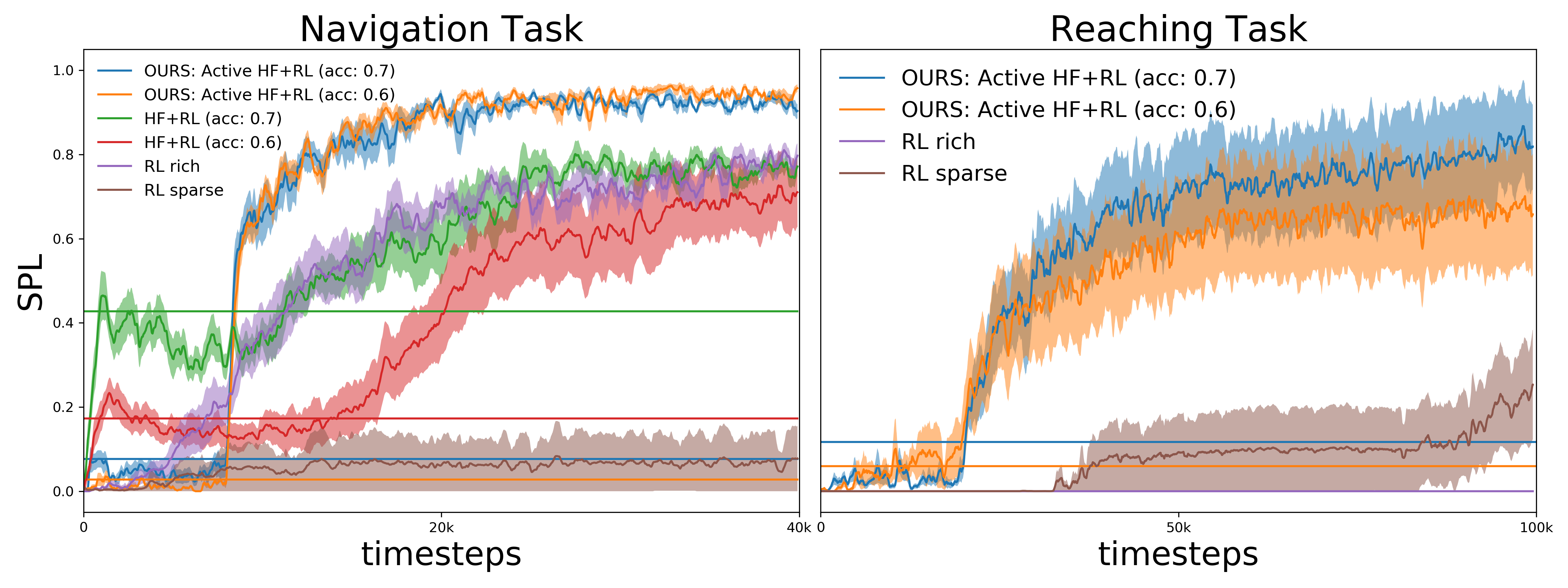}
\end{center}
\setlength{\belowcaptionskip}{-15pt}
\caption{\small Simulated Feedback Results.
Using the SPL metric, we compare the performance of our method (Active HF+RL) at two feedback accuracies (Blue: 0.7, Orange: 0.6) with RL sparse (Purple), RL rich (Brown). The plot shows the mean and 1/3 of the standard deviation over 10 different runs. We also show the average performance of $\pi_{HF}$ as blue and orange horizontal lines.
Despite its suboptimal performances, $\pi_{HF}$ accelerates reinforcement learning in sparse reward settings.
Meanwhile, RL with sparse/rich rewards struggles or even fails to learn the tasks. Notice that RL rich has SPL = 0 all the time for the 3D reaching task which has a high dimensional state space.
For the navigation task, we also show the result of a baseline method~\cite{akinola2019accelerated} (Green: $70\%$, Red: $60\%$) that does not use active learning and purified buffer. Active HF+RL outperforms this baseline with better RL policy after convergence.
}
\label{fig:sim_results}
% \vspace{-2mm}
\end{figure}
\begin{figure}
\vspace{2mm}
\begin{center}
    \includegraphics[width=1.0\linewidth]{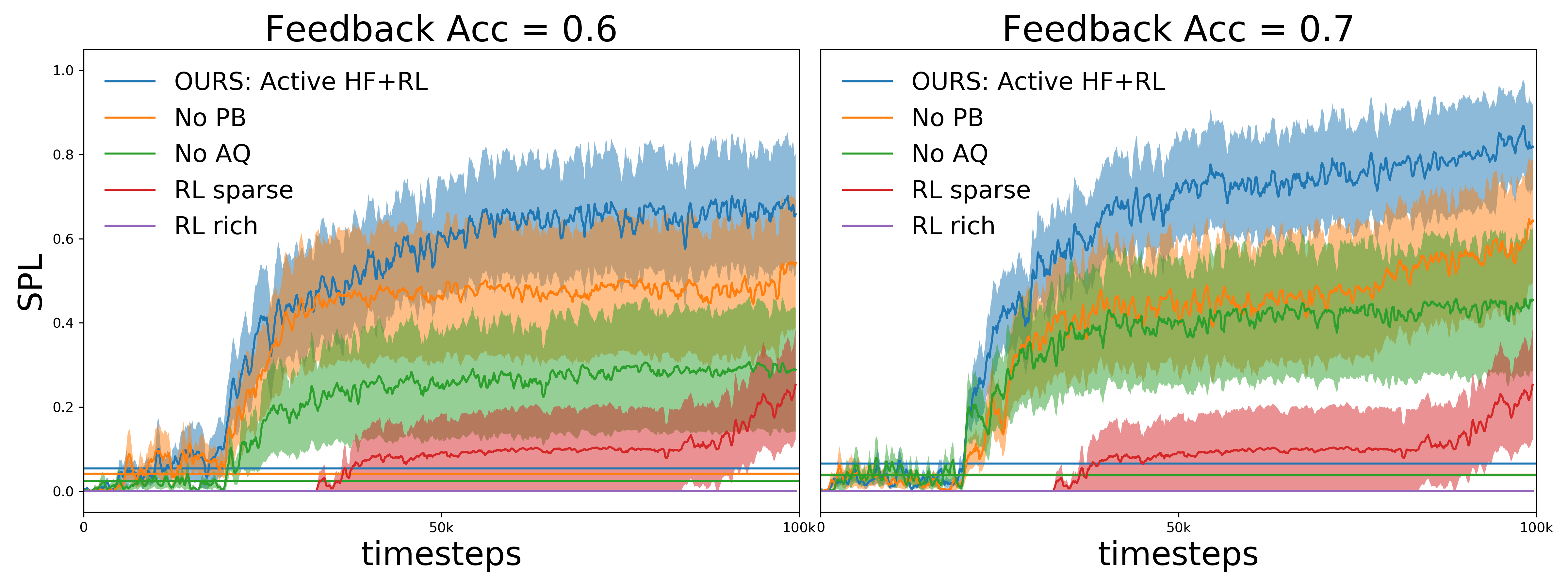}
\end{center}
\caption{\small Ablation Results. \textbf{Left}: $\mathcal{C} = 0.6$, \textbf{Right}: $\mathcal{C} = 0.7$.
We compare how removing Active Query (NO AQ) or Purified Buffer (No PB) affects the learning of $\pi_{HF}$ and $\pi_{RL}$. Again, the plot shows the mean and 1/3 of the standard deviation over 10 different runs, and the horizontal lines represent the average performance of $\pi_{HF}$ for each method. Without Active Query, we can see a $55\%/43\%$ drop in the $\pi_{HF}$ for $\mathcal{C} = 0.6/0.7$ respectively. This confirms our algorithm achieves same level of learning performance with less amount of the human feedback. Meanwhile, the $\pi_{HF}$ degraded by $23\%/40\%$ for "No PB".
}
\label{fig:ablation}
\vspace{-3mm}
\end{figure}

\begin{figure*}[h]
\vspace{2mm}
\begin{center}
    \includegraphics[width=1.0\linewidth]{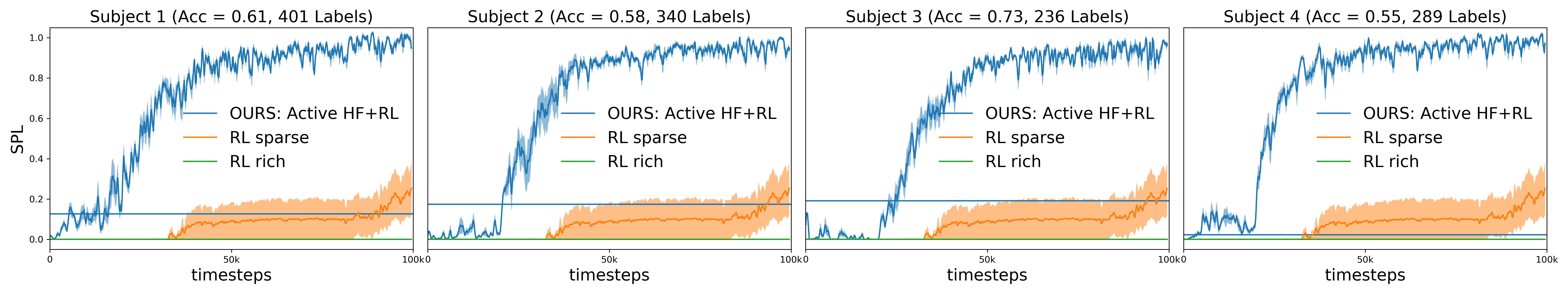}
\end{center}
\setlength{\belowcaptionskip}{-15pt}
\caption{\small Real Feedback Results for 4 Successful Subjects on a reaching task. Our method (Blue) uses feedback from human brain signals and accelerates RL learning.
Subject 4 has low $\pi_{HF}$ performance due to high feedback noise; our method still achieves optimal performance.}
\label{fig:real_results}
% \vspace{-3mm}
\end{figure*}

\section{Results}

With two challenging obstacle-avoidance tasks, we evaluate our proposed method.
To ensure repeatability, we first use simulated feedback from a scripted oracle. Then, we assess the performance of our system with real human feedback from 6 subjects.
% \todo{don't you use 6 later on...i am confused? Zizhao: Maybe modify after the real robot experiment}
For both experiments with simulated and real feedback, we compare our method (Active HF+RL) with RL algorithms learning from sparse/rich rewards (RL sparse/RL rich). We also compare with a baseline method~\cite{akinola2019accelerated} in the navigation task with simulated feedback.
We select the Success weighted by (normalized inverse) Path Length (SPL)~\cite{anderson2018evaluation} as the metric, which considers both success rate and path optimality.
For a fair comparison, we use the same architecture and hyperparameters for the RL part across all methods.
Moreover, RL sparse and RL rich also keep a buffer of their good experiences during training and use self-imitation learning component presented in our method.

\subsection{Learning from Simulated Feedback}
For simulated feedback, we test two accuracies: $\mathcal{C} = 0.6/0.7$. It evaluates how well the HF policy assists the RL learning with noisy feedback.
More consistent feedback requires fewer labels to learn a good $\pi_{HF}$, hence we query $1000/450$ feedback labels for $\mathcal{C} = 0.6/0.7$ respectively in the navigation tasks and $1000/300$ feedback labels in the reaching task.
% As it requires fewer labels to learn a good $\pi_{HF}$ from more consistent feedback, in the navigation task, for $\mathcal{C} = 0.6/0.7$, we query $1000/450$ feedback labels respectively. While in the reaching task, $1000/300$ feedback labels are queried respectively.
Meanwhile, we select the value of $\epsilon_{AQ} = 0.4/0.5, \epsilon_{PB} = 0.5/0.6$ using grid search.

Shown in Fig \ref{fig:sim_results}, most RL-sparse trials struggle to learn the task as the chance to reach the goal with random actions is very small. For RL-rich, even though it works well in the navigation task, in the harder reaching task, it is even worse than RL sparse as all trials fail as the method greedily maximize the rewards and cannot get over the obstacle. This suggests designing a successful dense reward is a nontrivial task and requires lots of trial and error. Instead, our method (Active HF+RL) solves the reaching task well for most trials. It confirms $\pi_{HF}$ still provides enough exploration to the goal and good experiences for $\pi_{RL}$ to learn from, even with feedback with significant noise.
Compared with a baseline method~\cite{akinola2019accelerated}, the imitation learning buffer enables the $\pi_{RL}$ to constantly review good exploration experiences from $\pi_{HF}$, and thus the RL agent can receive enough learning signals despite the poor performance of $\pi_{HF}$.

In Fig \ref{fig:ablation}, in the reaching task, we perform a series of ablation experiments to measure the importance of active feedback query and purified buffer when learning $\pi_{HF}$, as well as their effects on the training of $\pi_{RL}$.
Without the active query (\textbf{No AQ}), we can see a significant drop in the performance of $\pi_{HF}$ and $\pi_{RL}$. Especially, for $\mathcal{C} = 0.6$, the final performance of $\pi_{RL}$ is close to RL-sparse.
To see why, consider the case where the $\pi_{HF}$ agent already learned to move towards the goal but didn't know how to go over the obstacle yet. When collisions happen and the agent is reset to initial positions, the agent first needs to get close to the obstacle again. However, the "No AQ" will query feedback all the way even in well-learned states, which is a waste of the feedback. Besides, there can be wrong feedback due to noise, and it may harm or even destroy the learned policy. Instead, the "AQ" agent knows what it already knows and won't query feedback until it is near the obstacle. In this way, the queried feedback is more informative and the agent can learn to get over the obstacle more efficiently.
For agents without the purified buffer (\textbf{No PB}), there is also a drop in performance, especially for $\mathcal{C} = 0.7$. Meanwhile, we find feedback in the purified buffer has a much higher accuracy than the simulated accuracy, with $80\%$ compared with $60\%$, or $88\%$ compared with $70\%$. This affirms the purification buffer can assist $\pi_{HF}$ learning with more accurate labels. 

\subsection{Learning from Real Human Feedback}

We tested our Active HF+RL method on the reaching task with 6 human subjects providing feedback in the form of EEG signals while watching the simulated robot learn. First, the subject is trained for 5 minutes to get familiar with the paradigm and understand how the arm should move to the goal. Then, the subject has 20 1-min offline sessions to collect data for training the EEG classifier and finally provides feedback during 30 1-min online sessions to train the $\pi_{HF}$ policy. This $\pi_{HF}$ is subsequently used to guide the RL similar to the simulation experiments.
% Video of the experiments can be found at \url{http://crlab.cs.columbia.edu/brain_guided_rl/}.

Shown in Fig \ref{fig:real_results}, the low performing $\pi_{HF}$ policies from 4 subjects successfully guide the RL learning. For two other subjects, the EEG classifier accuracies are low ($\sim 51 \%$), and thus their feedback is too noisy to train a useful $\pi_{HF}$ to guide RL.
% \todo{how low is their accuracy - should we report this and also note that this is a limitation of this approach?}
A limitation of our approach is that we depend on the ability to detect error signals via EEG. Since detecting ErrP can vary across different individuals, our system does not work for individuals whose signals cannot be accurately classified.
Nevertheless, our extensive simulation experiments show that the proposed method handles significant feedback noise levels and can also be applied to other input modalities that have less variability across subjects.
\section{Discussion}
The experiments on navigation and reaching tasks with either simulated or real feedback show that our Active HF+RL can learn from feedback efficiently and accelerate RL in complex sparse reward environments. In contrast, RL learning from sparse or even rich rewards fails to finish the task. In the ablation studies, with the same amount of feedback, LfHF with Active Query learns much better human feedback policy and RL policy than the one without AQ. This confirms Active Query improves information gain from each query and significantly reduces the amount of expensive feedback needed. Meanwhile, when dealing with feedback inconsistency as large as $40\%$, the purified buffer can filter out noise and contain feedback of a much higher quality ($80\%$ accuracy). This guarantees our method is robust to low ErrP classification accuracy. Finally, even though $\pi_{HF}$ has suboptimal performance (e.g. subject 4), the imitation learning with Q-value filtering ensures RL will only learn from good guided explorations of $\pi_{HF}$ and repeatedly reviews them, ensuring RL can achieve optimal performance.
% \todo{did we scrap the physical experiments? Zizhao: We tried real robot experiments twice and will have a final try tomorrow. Depending on the results, we will decide to add it or not.}

\section{Conclusion}

This paper introduces Active Brain-Guided RL, a method to use human feedback evaluated from noisy and expensive EEG signals to bootstrap RL learning in sparse reward settings. We first train a HF policy with Active Query and Purified Buffer, and then the HF policy generates good experiences for RL to learn from. This method demonstrates robustness in three important ways: (1) When using Active Query to smartly select queries, it greatly improves feedback-efficiency and is robust to the limited amount of expensive human feedback. (2) With the Purfied Buffer, it filters out noise and learns from feedback of higher accuracy, making it robust to feedback inconsistency due to noisy EEG signals and poor classification accuracy. (3) It is also robust to the low success rate of the human feedback policy, since the human feedback policy still provides coarse guidance to the goal. Besides, the imitation learning with Q-value filter ensures the RL agent will constantly learn from good experiences of $\pi_{HF}$ and can outperform it when close to convergence. Different experiments using simulated or real feedback and corresponding ablation studies confirm our method can learn long-horizon tasks from sparse rewards with less amount of feedback than previous methods.



\section*{ACKNOWLEDGMENT}
We thank Prof. Paul Sadja and Pawan Lapborisuth for insightful conversations and advice in the course of the research. We also thank Kaveri Thakoor for helpful comments and feedback in the early draft of the paper.
We gratefully acknowledge Microsoft Inc. for their support of Iretiayo Akinola through the Microsoft Research PhD Fellowship Program.
% %% Use plainnat to work nicely with natbib. 

%%%%%%%%%%%%%%%%%%%%%%%%%%%%%%%%%%%%%%%%%%%%%%%%%%%%%%%%%%%%%%%%%%%%%%%%%%%%%%%%

\bibliographystyle{IEEEtran}
% \bibliography{IEEEabrv,IEEEexample}

% \nocite{*}
% \bibliographystyle{plainnat}
% \bibliography{references}

\end{document}